\documentclass{article}

\usepackage[final, nonatbib]{neurips_2024}
\usepackage[numbers]{natbib}
\usepackage{amsmath}
\usepackage{tikz}
\usetikzlibrary{arrows.meta, positioning}
\usepackage[utf8]{inputenc} 
\usepackage[T1]{fontenc}    
\usepackage{hyperref}       
\usepackage{url}            
\usepackage{booktabs}       
\usepackage{amsfonts}       
\usepackage{nicefrac}       
\usepackage{microtype}      
\usepackage{xcolor}
\usepackage{comment}
\usepackage{hyperref}  

\title{Advancing Agentic Systems: Dynamic Task Decomposition, Tool Integration and Evaluation using Novel Metrics and Dataset}

\author{
  Adrian Garrett Gabriel \\
  \And
  Alaa Alameer Ahmad \\
  \And
  Shankar Kumar Jeyakumar \\
  \AND
  CARIAD SE \\
  Major-Hirst-Straße 7 \\
  38442 Wolfsburg \\
  \texttt{marek.mayer@cariad.technology} \\
}

\begin{document}
\maketitle
\begin{abstract}
The rapid advancements in Large Language Models (LLMs) and their enhanced reasoning capabilities are opening new avenues for dynamic, context-aware task decomposition, and automated tool selection. These developments lay the groundwork for sophisticated autonomous agentic systems powered by LLMs, which hold significant potential for process automation across various industries. These systems demonstrate remarkable abilities in performing complex tasks, interacting with external systems to augment LLMs' knowledge, and executing actions autonomously. To address the challenges and harness the opportunities presented by these advances, this paper makes three key contributions.

\begin{itemize}
    \item We propose an advanced agentic framework designed to autonomously process multi-hop user queries by dynamically generating and executing task graphs, selecting appropriate tools, and adapting to real-time changes in task requirements or tool availability.
    \item We introduce novel evaluation metrics tailored for assessing agentic frameworks across diverse domains and tasks, namely Node F1 Score, Structural Similarity Index, and Tool F1 Score.
    \item We develop a specialized dataset based on the AsyncHow dataset to enable in-depth analysis of agentic behavior across varying task complexities.
\end{itemize}

Our findings demonstrate that asynchronous and dynamic task graph decomposition significantly improves system responsiveness and scalability, particularly in handling complex, multi-step tasks. Through detailed analysis, we observe that structural and node-level metrics are more critical in sequential tasks, whereas tool-related metrics dominate in parallel tasks. In particular, the Structural Similarity Index (SSI) emerged as the most significant predictor of performance in sequential tasks, while Tool F1 Score proved essential in parallel tasks. These findings highlight the need for balanced evaluation methods that capture both structural and operational aspects of agentic systems. Our specialized dataset enables comprehensive evaluation of these behaviors, providing valuable insights into improving overall system performance, with the importance of both structural and tool-related metrics validated through empirical analysis and statistical testing.

The evaluation of agentic systems presents unique challenges due to the intricate relationships between task execution, tool usage, and goal achievement. Our evaluation framework, validated through empirical analysis, offers valuable insights for improving the adaptability and reliability of agentic systems in dynamic environments.
\end{abstract}

\newpage
\section{Introduction}
Recent advances in LLMs have catalyzed the development of sophisticated agentic systems capable of automating multistep tasks, interacting with external systems, and adapting to changing contexts \cite{brown2020language, touvron2023llama}. These systems are promising in industries requiring autonomous workflow processing and tool integration. Despite their potential, LLM-based systems face limitations in industrial settings due to lack of training on proprietary data and the challenges of fine-tuning. Fine-tuning LLMs for each business use case requires costly, labor-intensive dataset collection and processing. In such contexts, LLMs are limited in their ability to manage real-time decision making in dynamic environments.

A scalable alternative to fine-tuning LLMs for each use case is the development of agentic systems that dynamically integrate external tools. Augmenting LLMs with tools allows these systems to handle complex queries and adapt without constant retraining. Agentic frameworks enable LLMs to decompose tasks into smaller sub-tasks, whose significance is highlighted in \cite{lin2024graphenhanced}, select the appropriate tools for each task, and adjust to real-time changes in tool availability. 

One of the first frameworks that allowed LLMs to interact with external tools is LangChain \href{https://www.langchain.com/}{Project}, which paved the way for more sophisticated agentic platforms such as BabyAGI \href{https://github.com/yoheinakajima/babyagi}{Project} and AutoGen \cite{wu2023autogenenablingnextgenllm}. These systems represent important steps toward the realization of autonomous AI agents, but are often constrained by high latency, limited adaptability, and insufficient support for dynamic tool integration. Moreover, current systems lack comprehensive evaluation methods that fully capture the complexity of task graph generation and tool selection, limiting their scalability and reliability in industrial applications.

To address these challenges, our work presents a framework\footnote{A detailed guide to replicate our agentic system is in Section \ref{supplementary_replicate_framework} of the Supplementary Material} that advances the capabilities of traditional agentic systems. Our framework integrates real-time tool selection, dynamic task graph generation, and an evaluation mechanism to assess agentic behavior across diverse tasks and domains. The proposed architecture consists of five core components: the Orchestrator that generates task graphs based on user queries, the Delegator that manages task distribution, ensuring seamless communication between tasks, the agents that autonomously execute tasks using LLMs, the tools that provide predefined functions necessary for task completion, and the Executor that handles the execution sequence, optimizing for both parallel and sequential task execution. 

A significant aspect of this work is the development of a comprehensive evaluation framework\footnote{A detailed guide to replicate our evaluation framework is in Section \ref{supplementary_replicate_evaluation} of the Supplementary Material}. Existing agentic systems often lack domain-specific metrics to rigorously assess their performance in handling task decomposition and tool integration. To fill this gap, we propose these novel metrics:

\textbf{Node and Tool F1 Scores}: These metrics assess the system’s precision and recall in matching task nodes to the expected task graph, ensuring accurate task decomposition, and in selecting the appropriate tools for each task within the graph.

\textbf{Structural Similarity Index (SSI)}: A metric that assesses the overall fidelity of the task graph generated by the system compared to the expected graph, capturing both node and edge similarities to ensure the system preserves the logical structure of tasks.

Additionally, we introduce a specialized dataset\footnote{The dataset is  available at this \href{https://anonymous.4open.science/r/AsyncHow-Based-Agentic-Systems-Evaluation-Dataset-C6BD/}{link}} designed to evaluate agentic systems, which allows a detailed analysis of the interdependencies between task decomposition, tool selection, and system performance, providing a foundation for evaluating agentic. 

In summary, this paper makes the following contributions:
\begin{itemize}
    \item An agentic framework for dynamic task decomposition, tool integration, and autonomous task execution, designed for complex, multi-hop queries.
    \item Novel evaluation metrics, including the Node F1 score, the Structural Similarity Index, and the Tool F1 score, for detailed task-specific assessment of agentic systems.
    \item A specialized dataset to evaluate agentic behavior across, supporting analysis of task graph generation, tool selection, and system performance.
\end{itemize}

The remainder of this paper is structured as follows - Section 2 discusses related work in agentic systems, task graph generation, and evaluation of agentic systems. Section 3 provides a detailed explanation of the architecture of our proposed framework. Section 4 presents our evaluation metrics and dataset, followed by empirical results in Section 5. Finally, Section 6 concludes with future directions and implications for agentic systems in real-world industrial settings.

\section{Related Work}
\subsection{Agentic Frameworks for Task Graph Generation and Tool Selection}
The field of AI agent systems powered by LLMs has seen substantial development, with various frameworks proposed to enhance collaboration, task automation, and system scalability. Recent surveys by Wang et al., Guo et al., Masterman et al., and Xi et al. have explored these advancements in detail, highlighting the evolving landscape of LLM-powered AI agents \cite{Wang_2024, guo2024largelanguagemodelbased, masterman2024landscapeemergingaiagent, xi2023risepotentiallargelanguage, du2024surveycontextawaremultiagentsystems}. In \cite{crawford2024bmw}, Crawford et al. thoroughly explored a flexible agent framework with a focus on the planning and execution components of an AI agent. \cite{lin2024graphenhanced} particularly explores the importance of planning and task graph generation.

Existing frameworks, such as LangGraph, AutoGen, and BabyAGI, provide various approaches to task generation, tool selection, and handling multi-hop user queries \cite{langgraph2024, chen2023autoagents, chen2023agentverse}. For instance, LangGraph enables stateful workflows with tasks managed cyclically or sequentially, but it does not dynamically adjust the task graph during execution as our proposed framework does \cite{bmw2024}. AutoGen and BabyAGI offer dynamic task generation but are limited by predefined toolsets and scalability issues, respectively \cite{autogen2024, autogen_no_code2023}. Our framework improves upon these methods by introducing real-time adaptability with a Task-Aware Semantic Tool Filtering mechanism, allowing for on-the-fly integration of new tools and dynamic task graph adjustments to handle increased task complexity \cite{dgnn_survey2024}. This capability ensures continuous and efficient task execution, enhancing existing approaches. Despite these advancements, several limitations and shortcomings exist in current frameworks:

\textbf{Lack of testability}: Existing frameworks often do not support rigorous unit testing of each component, making it challenging to ensure reliability and correctness in complex tasks \cite{autogen_no_code2023}. Our framework architecture is highly modular, enabling each agent to be tested individually, addressing this gap.

\textbf{High latency}: The absence of efficient task parallelization mechanisms leads to increased latency, hindering real-time performance \cite{autoagents2024, tdag2024}. Our system decomposes user queries into sub-tasks and executes them in parallel wherever possible, reducing latency.

\textbf{Limited customizability}: Many frameworks offer limited flexibility for customization, restricting their applicability across diverse domains and specific use cases \cite{autogen_no_code2023, autogen2024}. Our customizable design ensures applicability across various domains.

\subsection{Evaluation Metrics and Datasets for Agentic Frameworks}
Despite advancements in agentic frameworks, there remains a notable gap in comprehensive evaluation metrics and datasets that accurately assess the performance of these systems across diverse tasks and domains. Existing benchmarks, such as \textit{AgentBench} and \textit{VisualAgentBench}, focus on LLM performance across diverse environments and visual task automation but fall short in evaluating task graph structures in depth \cite{liu2023agentbench, liu2024visualagentbench}. Similarly, \textit{TASKBENCH} introduces "Tool Graphs" to measure task automation processes but provides limited analysis of intermediate steps \cite{shen2023taskbench}, while \textit{AgentQuest} emphasizes multi-step reasoning with little focus on task graph fidelity \cite{gioacchini2024agentquest}. Although \textit{VillagerAgent} explores multi-agent task dependencies in simulated environments, its focus is more on coordination than on detailed task decomposition \cite{dong2024villageragent}.

Our proposed evaluation framework fills these gaps by introducing detailed metrics—such as Node F1 Score, Structural Similarity Index, and Tool F1 Score—paired with specialized datasets to assess task decomposition, tool selection, and execution through task graph metrics. These metrics and datasets offer a granular analysis of agent performance, addressing the complexity of multi-step reasoning, task automation, and the interdependencies between evaluation metrics. This approach addresses gaps in current benchmarks by focusing on intermediate steps and structural fidelity, offering a more nuanced assessment applicable to real-world scenarios.

\section{Agentic Framework Architecture}
\begin{figure}[h!]
    \centering
    \includegraphics[width=0.9\linewidth]{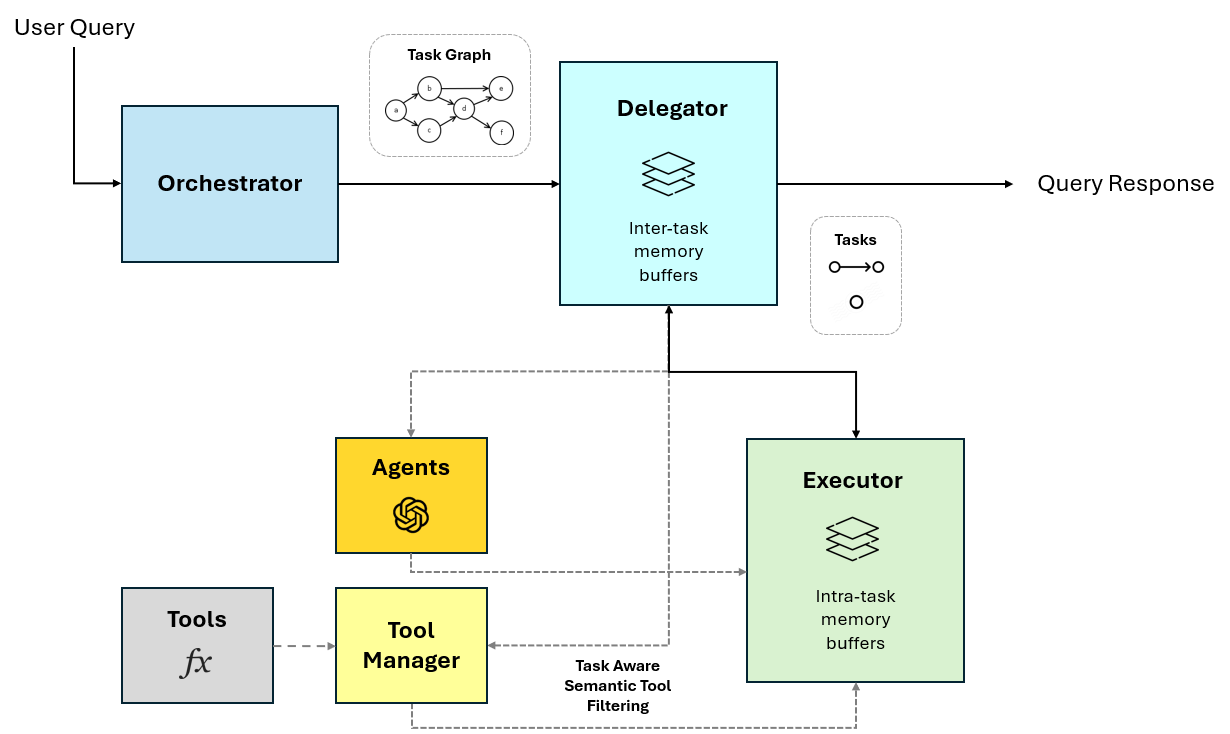}
    \caption{Overview of the framework architecture, showcasing the flow of a user query through the system.}
    \label{fig:pipeline-diagram}
\end{figure}

The framework consists of the following major components, as illustrated in Figure \ref{fig:pipeline-diagram}. These components include the Orchestrator, the Delegator, Tools, Agents, and the Executor. The diagram outlines the flow of a user query through these components, showing how each element interacts to produce a final response.

The Orchestrator as shown in Figure \ref{fig:pipeline-diagram} analyzes the user's input to produce a Directed Acyclic Graph (DAG) with task nodes and dependency edges. Asynchronous task graph decomposition, as explored by recent work on graph-enhanced LLMs, allows parallel task execution, enabling real-time adaptation to task changes and dependencies. This reduces execution time by shortening critical paths, which is crucial for handling complex, dynamic queries \cite{lin2024graphenhanced}. Hence, borrowing from this work and Graph Theory, the Orchestrator can be instructed to produce a task graph that is optimized for one or more of the following concepts:
\begin{itemize}
    \item \textbf{Coarse Grained Task Decomposition}: A strategy in which the user query is decomposed into a relatively small number of large tasks, each representing a substantial amount of work. This approach minimizes the overhead associated with task management, such as scheduling, synchronization, and inter-task communication, by focusing on larger, more independent units of work \cite{foster1995designing}, \cite{quinn2004parallel}. While this can reduce the potential for parallelism, it simplifies execution and can be more efficient in scenarios where the overhead of managing numerous small tasks is prohibitive.
    \item \textbf{Fine Grained Task Decomposition}: A strategy where a user query is broken down into a large number of small, granular tasks. Each task represents a minimal unit of work, allowing for a high degree of parallelism as many tasks can be executed simultaneously. However, this approach often requires more sophisticated management of task dependencies, communication, and synchronization, which can introduce significant overhead \cite{foster1995designing}, \cite{quinn2004parallel}.
    \item \textbf{Critical Path Optimization}: A strategy that focuses on identifying and shortening the critical path—the longest sequence of dependent tasks that determines the minimum time required to obtain a complete response to a User Query. By optimizing tasks in this path, the overall latency can be reduced \cite{kerzner2017project}.
\end{itemize}

As depicted in Figure \ref{fig:pipeline-diagram}, the Delegator receives the task graph from the Orchestrator and is responsible for assigning tasks to the appropriate agents or tools. It plays a critical role in managing intra-task and inter-task communication by utilizing inter-task memory buffers, ensuring that each task has the necessary context and data from its dependent predecessors.

The Delegator also consolidates the results from all completed tasks to form the final response to the user query. In scenarios where tasks have direct or indirect dependencies, the Delegator ensures that the inter-task memory buffers are populated with the necessary task descriptions and execution results, thereby enabling subsequent tasks to execute with full awareness of the context provided by their dependencies.

Within the framework, agents are dynamic components, as illustrated in the diagram (Figure \ref{fig:pipeline-diagram}). They are responsible for utilizing a LLM to execute specific tasks. For instance, the PythonAgent can generate Python functions on the fly, acting as tools that can be called with arguments. The PythonAgent is particularly useful when a task requires a function that is not yet available in the existing set of tools.

Tools are static components that consist of pre-defined Python functions, which can accept arguments and execute code as needed for various tasks. As shown in Figure \ref{fig:pipeline-diagram}, the tool manager is responsible for task-aware semantic tool filtering. This ensures that only the most relevant tools are passed to the LLM based on the specific task requirements, rather than overwhelming it with the entire tool cache. This semantic filtering enhances the efficiency and accuracy of task execution by reducing unnecessary processing overhead \cite{crawford2024bmw}.

\begin{figure}[h]
    \centering
    \includegraphics[width=.9\linewidth]{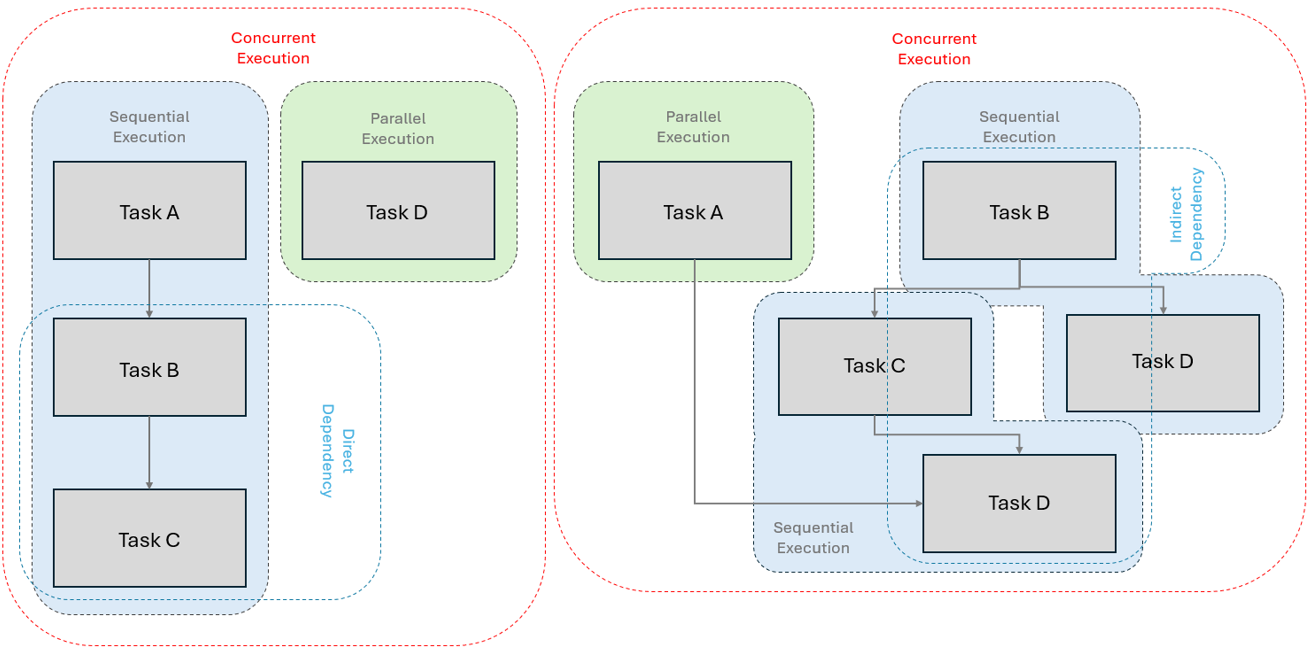}
    \caption{Execution flow managed by the Executor showcasing a simpler task flow on the left and a more complex flow on the right.}
    \label{fig:executor-diagram}
\end{figure}

The Executor (see Figure \ref{fig:executor-diagram}) is responsible for carrying out the execution of the tasks specified in the Task Graph produced by the Orchestrator. The Executor interacts with various components such as the Delegator, Agents, and Tool Manager, managing both intra-task and inter-task memory buffers to ensure efficient task execution. It processes tasks while respecting both direct and indirect dependencies within the task graph, optimizing for parallel execution where possible while ensuring the correct task order when dependencies exist.

In terms of execution, tasks that are independent of each other can be run concurrently, enabling parallelism where possible to maximize efficiency, an example of which is shown in Appendix \ref{appendix_additional_executor_plots}. For tasks with direct dependencies, sequential execution is required to ensure that input data from predecessor tasks is available before execution proceeds. Indirect dependencies, where tasks depend on the results of other tasks via intermediates, are also managed by the system to enforce the correct execution order.

The execution flow in Figure \ref{fig:executor-diagram} illustrates two typical scenarios. On the left side, a simpler case is presented where tasks in a task graph are grouped and executed sequentially (Task A $\rightarrow$ Task B $\rightarrow$ Task C) with direct dependencies, while Task D is executed in parallel. On the right side, a more complex scenario shows Task A being executed in parallel, while Tasks B, C, and D have both direct and indirect dependencies, requiring a combination of sequential and parallel execution.

\section{Evaluation Framework}
Evaluating agentic systems necessitates a multidimensional analysis of all the individual components as well as the final output. Our approach focuses on evaluating intermediate steps as well as final outcomes, using a robust set of metrics designed to assess the system's ability to decompose tasks, select appropriate tools, and execute tasks effectively. This evaluation framework is grounded in existing literature, including the gaps identified by Gioacchini et al.\cite{gioacchini2024agentquest}, Shen et al.\cite{shen2023taskbench}, and Liu et al.\cite{liu2023agentbench}, which highlight the need for a more granular evaluation of agentic behaviors in LLMs.
To address these gaps, we developed a comprehensive evaluation dataset inspired by Lin et al.\cite{lin2024graph}, integrating task graphs, tool usage and final outputs. This dataset supports a more nuanced analysis of agentic systems by capturing both intermediate processes and final results, ensuring a holistic evaluation of system performance.

\subsection{Dataset Creation}
Our dataset is specifically designed to evaluate the agentic behavior of LLM-driven systems across various domains and tasks. The dataset creation process was structured to ensure representativeness, and relevance to real-world scenarios based on the \textbf{AsyncHow} \cite{lin2024graph} dataset. 
The \textbf{AsyncHow} dataset was particularly suited for this study due to its unique structure that covers parallel, and sequential task graphs. This comprehensive coverage allows for a thorough evaluation of agentic systems, which need to handle different types of task relationships and dependencies. The dataset's design, which includes a mix of simple linear workflows and more intricate interdependent tasks, supports the comprehensive evaluation of the system's ability to manage these varying complexities effectively. The \textbf{AsyncHow} dataset's validation for each scenario within the parallel and sequential categories of task graphs ensures that it is a robust foundation for evaluating the proposed agentic systems. As this is one of the most important steps in an agentic system we found it ideal to use it as a foundational dataset. Below is a detailed breakdown of the steps involved.

\textbf{Task Graph Construction}: We randomly sampled 50 task graphs from the \textbf{AsyncHow} dataset, ensuring diversity while maintaining empirical manageability for the evaluation of agentic systems. By selecting 50 scenarios for the parallel and sequential task graph category, the study ensured that the dataset remains robust yet feasible for in-depth analysis. The empirical soundness of selecting 50 scenarios, resulting in more than 250 tools, was determined to be sufficient for evaluating the correctness of tool selection and the overall performance of the agentic systems.

\textbf{Tool Function Generation:}
Tool descriptions were parsed and translated into synthetic Python functions. These functions were designed to replicate the behavior of real-world tools, ensuring that the agent's interactions with these tools are realistic and contextually appropriate. These functions contribute to creating a realistic and challenging environment for the agentic system as they span a variety of tasks, such as simulating the return of data as JSON from API calls or the data from a candidate for a potential job interview. The functions were then executed to generate final responses for each scenario, providing a benchmark for evaluating the system's performance.

\textbf{Final Dataset Composition:}
The final dataset includes a comprehensive set of components: scenario names, task graphs, tool functions, expected tool call sequences, gold standard responses, and complexity categories for each scenario. This structure facilitates a detailed evaluation of both intermediate steps and final outcomes, enabling a more nuanced assessment of the agent's performance. 

With this data, we can evaluate each component of our framework: 
\begin{itemize}
    \item \textbf{Task graph composition:} We can compare the task graph from the evaluation dataset to the task graph of our agentic system and validate it according to several similarity metrics (\ref{eval_metrics}).
    \item \textbf{Tool selection:} We can match the list of expected tools selected against the list of tools our agentic system chose.
    \item \textbf{Answer generation:} We can judge the gold standard answer against the answer the agentic system created.
\end{itemize}

\subsection{Evaluation Metrics} \label{eval_metrics}
To rigorously evaluate the agentic system's performance, we employ a set of metrics that measure the accuracy and effectiveness of task decomposition, tool selection, and task execution. These metrics provide a comprehensive assessment of the system's capabilities, ensuring a thorough evaluation of task graphs, tool identification, and answer generation accuracy.

\textbf{Precision, Recall, and F1 Score} are core metrics used to evaluate the accuracy and completeness of tool identification. The precision score reflects the system's ability to correctly identify relevant tools without including false positives, while recall measures its success in identifying all relevant elements. The F1 Score balances precision and recall, providing a comprehensive view of the system’s performance in identifying tools. 

\begin{center}
\scriptsize 
\begin{align*}
    \text{Precision}_{\text{tool}} &= \frac{TP_{\text{tool}}}{TP_{\text{tool}} + FP_{\text{tool}}}, \quad
    \text{Recall}_{\text{tool}} = \frac{TP_{\text{tool}}}{TP_{\text{tool}} + FN_{\text{tool}}}, \quad
    \text{F1 Score}_{\text{tool}} = \frac{2 \times \text{Precision}_{\text{tool}} \times \text{Recall}_{\text{tool}}}{\text{Precision}_{\text{tool}} + \text{Recall}_{\text{tool}}}
\end{align*}
\end{center}

\begin{center}
\scriptsize 
\begin{align*}
    \text{Precision}_{\text{node}} &= \frac{TP_{\text{node}}}{TP_{\text{node}} + FP_{\text{node}}}, \quad
    \text{Recall}_{\text{node}} = \frac{TP_{\text{node}}}{TP_{\text{node}} + FN_{\text{node}}}, \quad
    \text{F1 Score}_{\text{node}} = \frac{2 \times \text{Precision}_{\text{node}} \times \text{Recall}_{\text{node}}}{\text{Precision}_{\text{node}} + \text{Recall}_{\text{node}}}
\end{align*}
\end{center}

\begin{center}
\scriptsize 
\begin{align*}
    \text{Precision}_{\text{edge}} &= \frac{TP_{\text{edge}}}{TP_{\text{edge}} + FP_{\text{edge}}}, \quad
    \text{Recall}_{\text{edge}} = \frac{TP_{\text{edge}}}{TP_{\text{edge}} + FN_{\text{edge}}}, \quad
    \text{F1 Score}_{\text{edge}} = \frac{2 \times \text{Precision}_{\text{edge}} \times \text{Recall}_{\text{edge}}}{\text{Precision}_{\text{edge}} + \text{Recall}_{\text{edge}}}
\end{align*}
\end{center}

These formulas are applicable for tool identification, node and edge matching, as accurate identification is crucial for the successful decomposition and execution of tasks by the agent.

\textbf{Node and Edge Matching:} We also assess how well the system matches nodes (tasks) and edges (dependencies) in the task graph to the expected structure. These metrics are key to evaluating the structural integrity and correctness of task decomposition.

\textbf{Node Label Similarity:} To assess how semantically similar the nodes in the actual task graph are to those in the expected graph, we use a cosine similarity measure based on node labels:
\begin{equation}
    \text{Node Label Similarity} = \frac{1}{|N_{\text{expected}}|} \sum_{i=1}^{|N_{\text{expected}}|} \max_{j \in N_{\text{actual}}} \text{CosineSim}(L_i, L_j)
\end{equation}

\textbf{Graph Evaluation:} Beyond node and edge matching, we employ metrics that evaluate the entire structure of the task graph, providing a more holistic assessment of its accuracy.

\begin{itemize}
    \item \textbf{Graph Edit Distance (GED):} GED quantifies the dissimilarity between the actual task graph and the expected graph by calculating the minimum number of edit operations (insertions, deletions, substitutions) needed to transform one graph into the other. It provides a more granular view of graph differences than simple node or edge matching.
    \item \textbf{Structural Similarity Index (SSI):} This metric combines node and edge similarities into a single score, offering a comprehensive evaluation of the task graph’s structural fidelity:
    \begin{equation}
    \text{SSI} = \frac{\text{Node Label Similarity} + \text{Edge F1 Score}}{2}
    \end{equation}
    \item \textbf{Path Length Similarity:} This metric compares the lengths of paths between corresponding nodes in both the actual and expected task graphs, providing insight into how well the system captures the broader structure of task dependencies:
    \begin{equation}
    S_{PL} = \frac{1}{|V|^2} \sum_{(u, v) \in V_1 \times V_2} \exp\left(-\alpha \left|d_{G_1}(u, v) - d_{G_2}(u, v)\right|\right)
    \end{equation}
\end{itemize}

\textbf{Complexity Score:} The complexity score as defined by \cite{lin2024graph} as the total number of vertices and edges within a task graph: $|V| + |E|$, where $V$ is the number of nodes and $E$ is the number of edges. This metric provides a measure of the graph's structural complexity and helps compare task graphs of varying sizes.

\newpage
\section{Results and Discussion}
Upon investigating our agentic framework we noticed that explicitly mentioning decomposition strategies (coarse grained vs fine grained) allowed the system to adapt based on the complexity and interdependence of tasks extracted from a user query and saw a significant improvement in task accuracy and reduction in inefficiency - lesser amount of redundant tasks were produced. This is in agreement with \cite{cao2021coarse, liu2020fine}, which highlight the importance of multi-granularity approaches in AI frameworks, particularly in complex domains like multi-hop question answering.

We conducted an in-depth analysis of the proposed metrics on our agentic system to understand their impact on the system's performance for both sequential and parallel tasks. The analysis revealed that structural and node-level metrics are more critical in sequential tasks, while tool-related metrics are prominent in parallel tasks. Our choice of evaluation metrics is validated by prior research on LLMs and their performance in graph-based tasks. Studies have shown that combining precision and recall, such as in the Node F1 Score, is essential for accurately evaluating systems that interact with complex graph structures, as it minimizes false positives and negatives \cite{wang2024microstructures}. The Structural Similarity Index (SSI) has also been validated as a reliable metric for assessing the preservation of both the functional and structural aspects of task graphs \cite{wills2020metrics}. 

The \textbf{Structural Similarity Index} (SSI) emerged as the most significant predictor of Answer Score, with a strong positive correlation (r = 0.470, p < 0.001). Plot can be seen in the Appendix in Figure \ref{fig:seq_correlations}. Node Label Similarity also showed a substantial positive correlation (r = 0.447, p < 0.01). Interestingly, Expected Task Complexity exhibited a moderate negative correlation (r = -0.293, p < 0.05), suggesting that as tasks become more complex, performance tends to decrease. These correlations were further reflected in real-world applications of the agentic system, where scenarios requiring fine-grained task decomposition revealed high Node Precision and Recall, but also highlighted challenges with managing complex dependencies, as indicated by lower Edge F1 Scores.
When considering the absolute coefficients from our linear regression model, SSI again emerged as the most important feature, followed by Edge F1 Score and Node Label Similarity. This aligns with our correlation analysis, emphasizing the importance of structural accuracy in sequential tasks.
The model achieved an R-squared value of 0.3631, indicating that these features explain approximately 36.31\% of the variance in Answer Score for \textbf{sequential} tasks.

For parallel tasks, we observed a shift in the importance of metrics, with tool-related features gaining prominence. Plots can be seen in Appendix \ref{appendix_eval_plots} in Figure \ref{fig:para_correlations}. Tool F1 Score showed the strongest correlation with Answer Score (r = 0.476, p < 0.001), closely followed by Tool Recall (r = 0.474, p < 0.01) and Tool Precision (r = 0.414, p < 0.01). SSI and Node Label Similarity both demonstrated moderate positive correlations (r = 0.380, p < 0.05 for both). Interestingly, our regression analysis revealed that SSI and Node Label Similarity had the highest importance scores, despite not having the strongest correlations. This suggests a complex interplay between these structural features and tool-related metrics in parallel tasks. The model for parallel tasks achieved an R-squared value of 0.3933, explaining 39.33\% of the variance in Answer Score.

Our findings highlight differences in the factors influencing agentic system performance between sequential and parallel tasks. \textbf{Structural metrics} (SSI and Node Label Similarity) are important across both task types, but their relative importance is higher in sequential tasks. Practical application of our evaluation framework also supports these findings, where task decomposition in real-world scenarios displayed strong structural performance but revealed weaknesses in dependency management (as seen in lower Edge F1 Scores). \textbf{Tool-related metrics} (Tool F1 Score, Recall, and Precision) are more strongly correlated with performance in parallel tasks, suggesting that effective tool selection and usage become critical in more complex, non-linear task structures. \textbf{Edge-related metrics} (Edge Precision and Edge F1 Score) show some importance in sequential tasks but are not significant in parallel tasks, possibly due to the more complex relationships between nodes in parallel structures.

\section{Limitations}
While the proposed agentic framework offers significant advancements in evaluating agentic systems, it faces several limitations. A primary issue is the lack of support for multi-agent communication, which limits the framework’s effectiveness in scenarios requiring task coordination among multiple agents. The system is only suited for single-agent environments or cases where agents operate independently. Expanding this capability would significantly enhance its versatility in complex settings.

Precision, Recall, and F1 Score—are also sensitive to dataset imbalances. When there are very few or many expected tools, these metrics may not capture performance nuances effectively, leading to skewed results. Another limitation is that these metrics do not account for the context in which the tools are applied. A tool may be identified correctly, but if used in an inappropriate context, this error is not reflected in the metrics, potentially leading to inaccurate performance assessments. Calculating Graph Edit Distance (GED) is computationally intensive, especially for large graphs, and this sensitivity of GED to cost, along with the ambiguity in its interpretation, reduces its effectiveness in evaluating complex task graphs. The matching of nodes and edges can suffer when multiple nodes or edges possess similar labels, which can require manual intervention or the implementation of additional rules to resolve conflicts. This increases the complexity of the task and reduces the reliability of fully automated methods. Moreover, edge matching is heavily dependent on the accuracy of node matching, which means that any errors in node similarity evaluations can propagate and negatively affect the assessment of task dependencies. The Structural Similarity Index (SSI) also presents certain drawbacks. First, it is highly sensitive to node similarity scores, which can be problematic when nodes - representing tasks - are described in significantly different ways. This can distort the overall similarity measure. Additionally, SSI assigns equal weight to node and edge similarities, which may not be appropriate in all cases. In some situations, task dependencies, represented by edges, may be more important than the individual tasks themselves, and equal weighting could overlook this context. Finally, SSI computation, particularly for large graphs can be computationally expensive, further complicating its application in large-scale evaluations.

\section{Future Work}
Building on the limitations identified, future research should focus on the following. First, the integration of multi-agent communication protocols would allow for dynamic collaboration, enabling the system to handle more complex, real-time scenarios that require the collaboration of multiple agents. Additionally, introducing causal inference methods to the evaluation framework would offer deeper insights into system performance by moving beyond correlation-based metrics. Optimizing scalability remains a critical challenge, especially for large-scale real-time environments, where reducing latency and computational overhead is essential. Finally, developing more sophisticated datasets that reflect real-world randomness and multi-agent interactions is crucial for testing and improving the scalability of agentic systems. Refer to Appendix \ref{appendix_additional_future_work} for detailed concrete future research directions to extend this work.

\section{Conclusion}
Analysis of our agentic framework revealed that task graph based decomposition \cite{lin2024graph} and explicitly integrating coarse-grained and fine-grained decomposition strategies led to improved task accuracy and a reduction in inefficiencies by minimizing redundant tasks. This adaptability, driven by the complexity and interdependence of user query tasks, highlights the critical role of multi-granularity approaches in agentic systems. Moreover, while these strategies enhanced system performance, challenges in managing complex dependencies, especially in fine-grained tasks, became evident. 

Our work also presents a comprehensive framework for evaluating agentic systems driven by LLMs. We introduce proper metrics, including the \textbf{Node F1 Score}, \textbf{Structural Similarity Index}, and \textbf{Tool F1 Score}, to assess the performance of these systems in task decomposition and tool integration scenarios. Additionally, our specialized dataset created for this study enables an in-depth analysis of agentic behavior across various task complexities. Our findings emphasize the importance of both \textbf{structural and tool-related metrics} in determining overall system performance, with notable differences between sequential and parallel tasks. While structural metrics are more critical in sequential tasks, tool usage becomes more significant in parallel tasks, highlighting the need for balanced evaluation methods that capture both the structural and operational aspects of agentic systems. The proposed framework and evaluation methodology provide a strong foundation for further research. Addressing the outlined limitations and pursuing the proposed future work will be critical in advancing agentic systems capable of handling more complex, real-time, and collaborative environments.

\bibliography{bibliography}

\begin{thebibliography}{10}

\bibitem{brown2020language}
T.~B. Brown, B.~Mann, N.~Ryder, M.~Subbiah, J.~Kaplan, P.~Dhariwal, A.~Neelakantan, P.~Shyam, G.~Sastry, A.~Askell, et~al.
\newblock Language models are few-shot learners.
\newblock {\em arXiv preprint arXiv:2005.14165}, 2020.

\bibitem{cao2021coarse}
X.~Cao and Y.~Liu.
\newblock Coarse-grained decomposition and fine-grained interaction for multi-hop question answering.
\newblock {\em arXiv preprint arXiv:2101.05988}, 2021.

\bibitem{chen2023autoagents}
G.~Chen, S.~Dong, Y.~Shu, G.~Zhang, J.~Sesay, B.~F. Karlsson, J.~Fu, and Y.~Shi.
\newblock Autoagents: A framework for automatic agent generation.
\newblock {\em arXiv preprint arXiv:2309.17288}, 2023.

\bibitem{chen2023agentverse}
W.~Chen, Y.~Su, J.~Zuo, C.~Yang, C.~Yuan, C.-M. Chan, H.~Yu, Y.~Lu, Y.-H. Hung, C.~Qian, et~al.
\newblock Agentverse: Facilitating multi-agent collaboration and exploring emergent behaviors.
\newblock In {\em The Twelfth International Conference on Learning Representations}, 2023.

\bibitem{crawford2024bmw}
N.~Crawford, E.~B. Duffy, I.~Evazzade, T.~Foehr, G.~Robbins, D.~K. Saha, J.~Varma, and M.~Ziolkowski.
\newblock Bmw agents--a framework for task automation through multi-agent collaboration.
\newblock {\em arXiv preprint arXiv:2406.20041}, 2024.

\bibitem{bmw2024}
N.~Crawford et~al.
\newblock Bmw agents: A framework for task automation through multi-agent collaboration.
\newblock {\em arXiv preprint arXiv:2406.20041}, 2024.

\bibitem{dong2024villageragent}
Y.~Dong, X.~Zhu, Z.~Pan, L.~Zhu, and Y.~Yang.
\newblock Villageragent: A graph-based multi-agent framework for coordinating complex task dependencies in minecraft.
\newblock {\em arXiv preprint arXiv:2406.05720}, 2024.

\bibitem{du2024surveycontextawaremultiagentsystems}
H.~Du, S.~Thudumu, R.~Vasa, and K.~Mouzakis.
\newblock A survey on context-aware multi-agent systems: Techniques, challenges and future directions, 2024.

\bibitem{foster1995designing}
I.~Foster.
\newblock {\em Designing and Building Parallel Programs: Concepts and Tools for Parallel Software Engineering}.
\newblock Addison-Wesley, 1995.

\bibitem{gioacchini2024agentquest}
L.~Gioacchini, G.~Siracusano, D.~Sanvito, K.~Gashteovski, D.~Friede, R.~Bifulco, and C.~Lawrence.
\newblock Agentquest: A modular benchmark framework to measure progress and improve llm agents.
\newblock {\em arXiv preprint arXiv:2404.06411}, 2024.

\bibitem{guo2024largelanguagemodelbased}
T.~Guo, X.~Chen, Y.~Wang, R.~Chang, S.~Pei, N.~V. Chawla, O.~Wiest, and X.~Zhang.
\newblock Large language model based multi-agents: A survey of progress and challenges, 2024.

\bibitem{autoagents2024}
A.~Gupta et~al.
\newblock Autoagents: A framework for task generation and tool selection in multi-agent systems.
\newblock {\em arXiv preprint arXiv:2309.17288}, 2024.

\bibitem{dgnn_survey2024}
X.~Jin et~al.
\newblock A comprehensive survey of dynamic graph neural networks: Models, frameworks, benchmarks, experiments and challenges.
\newblock {\em arXiv preprint arXiv:2405.00476}, 2024.

\bibitem{kerzner2017project}
H.~Kerzner.
\newblock {\em Project Management: A Systems Approach to Planning, Scheduling, and Controlling}.
\newblock Wiley, 2017.

\bibitem{langgraph2024}
LangGraph.
\newblock Langgraph: Build resilient language agents as graphs.
\newblock \url{https://github.com/langchain-ai/langgraph}, 2024.

\bibitem{lin2024graphenhanced}
F.~Lin et~al.
\newblock Graph-enhanced large language models in asynchronous plan reasoning.
\newblock {\em arXiv preprint arXiv:2402.02805}, 2024.

\bibitem{lin2024graph}
F.~Lin, E.~La~Malfa, V.~Hofmann, E.~M. Yang, A.~Cohn, and J.~B. Pierrehumbert.
\newblock Graph-enhanced large language models in asynchronous plan reasoning.
\newblock {\em arXiv preprint arXiv:2402.02805}, 2024.

\bibitem{liu2020fine}
X.~V. Lin, R.~Socher, and C.~Xiong.
\newblock Multi-hop knowledge graph reasoning with reward shaping.
\newblock {\em Proceedings of the 2018 Conference on Empirical Methods in Natural Language Processing (EMNLP)}, pages 3243--3253, 2018.

\bibitem{liu2023agentbench}
X.~Liu, H.~Yu, H.~Zhang, Y.~Xu, X.~Lei, H.~Lai, Y.~Gu, H.~Ding, K.~Men, K.~Yang, et~al.
\newblock Agentbench: Evaluating llms as agents.
\newblock {\em arXiv preprint arXiv:2308.03688}, 2023.

\bibitem{liu2024visualagentbench}
X.~Liu, T.~Zhang, Y.~Gu, I.~L. Iong, Y.~Xu, X.~Song, S.~Zhang, H.~Lai, X.~Liu, H.~Zhao, et~al.
\newblock Visualagentbench: Towards large multimodal models as visual foundation agents.
\newblock {\em arXiv preprint arXiv:2408.06327}, 2024.

\bibitem{masterman2024landscapeemergingaiagent}
T.~Masterman, S.~Besen, M.~Sawtell, and A.~Chao.
\newblock The landscape of emerging ai agent architectures for reasoning, planning, and tool calling: A survey, 2024.

\bibitem{quinn2004parallel}
M.~J. Quinn.
\newblock {\em Parallel Programming in C with MPI and OpenMP}.
\newblock McGraw-Hill, 2004.

\bibitem{autogen_no_code2023}
M.~Research.
\newblock Autogen: A no-code developer tool for building and debugging multi-agent systems.
\newblock {\em arXiv preprint arXiv:2408.15247}, 2023.

\bibitem{shen2023taskbench}
Y.~Shen, K.~Song, X.~Tan, W.~Zhang, K.~Ren, S.~Yuan, W.~Lu, D.~Li, and Y.~Zhuang.
\newblock Taskbench: Benchmarking large language models for task automation.
\newblock {\em arXiv preprint arXiv:2311.18760}, 2023.

\bibitem{touvron2023llama}
H.~Touvron, T.~Lavril, G.~Izacard, X.~Martinet, M.-A. Lachaux, T.~Lacroix, B.~Rozi\`{e}re, N.~Goyal, E.~Hambro, F.~Azhar, et~al.
\newblock Llama: Open and efficient foundation language models.
\newblock {\em arXiv preprint arXiv:2302.13971}, 2023.

\bibitem{Wang_2024}
L.~Wang, C.~Ma, X.~Feng, Z.~Zhang, H.~Yang, J.~Zhang, Z.~Chen, J.~Tang, X.~Chen, Y.~Lin, W.~X. Zhao, Z.~Wei, and J.~Wen.
\newblock A survey on large language model based autonomous agents.
\newblock {\em Frontiers of Computer Science}, 18(6), 2024.

\bibitem{wang2024microstructures}
Y.~Wang, H.~Cui, and J.~Kleinberg.
\newblock Microstructures and accuracy of graph recall by large language models.
\newblock {\em arXiv preprint arXiv:2402.11821}, 2024.

\bibitem{wills2020metrics}
P.~Wills and F.~G. Meyer.
\newblock Metrics for graph comparison: a practitioner’s guide.
\newblock {\em Plos one}, 15(2):e0228728, 2020.

\bibitem{wu2023autogenenablingnextgenllm}
Q.~Wu, G.~Bansal, J.~Zhang, Y.~Wu, B.~Li, E.~Zhu, L.~Jiang, X.~Zhang, S.~Zhang, J.~Liu, A.~H. Awadallah, R.~W. White, D.~Burger, and C.~Wang.
\newblock Autogen: Enabling next-gen llm applications via multi-agent conversation, 2023.

\bibitem{autogen2024}
Q.~Wu et~al.
\newblock Autogen: Enabling next-generation large language model applications.
\newblock {\em Microsoft Research}, 2024.

\bibitem{xi2023risepotentiallargelanguage}
Z.~Xi, W.~Chen, X.~Guo, W.~He, Y.~Ding, B.~Hong, M.~Zhang, J.~Wang, S.~Jin, E.~Zhou, R.~Zheng, X.~Fan, X.~Wang, L.~Xiong, Y.~Zhou, W.~Wang, C.~Jiang, Y.~Zou, X.~Liu, Z.~Yin, S.~Dou, R.~Weng, W.~Cheng, Q.~Zhang, W.~Qin, Y.~Zheng, X.~Qiu, X.~Huang, and T.~Gui.
\newblock The rise and potential of large language model based agents: A survey, 2023.

\bibitem{tdag2024}
L.~Zhang et~al.
\newblock Tdag: A multi-agent framework based on dynamic task decomposition and agent generation.
\newblock {\em arXiv preprint arXiv:2402.10178}, 2024.

\end{thebibliography}
\newpage
\section*{Supplementary Material}

\renewcommand{\thesection}{S\arabic{section}} 
\setcounter{section}{1}  

\subsection{Replicating the Agentic Task Decomposition Framework}\label{supplementary_replicate_framework}

This section provides a detailed guide for replicating the agentic task decomposition framework, including Python function shells with brief descriptions. This guide assumes familiarity with Python, machine learning, and related libraries.

Our framework decomposes user queries into a Directed Acyclic Graph (DAG) of tasks and uses an LLM for tool selection and execution. The system components include the \textbf{Orchestrator}, \textbf{Delegator}, \textbf{ToolManager}, and \textbf{GraphExecutor}. These work together to generate tasks, select tools, and ensure parallel or sequential execution.

\subsubsection{Prerequisites}

The following Python libraries are required (from \texttt{requirements.txt}):
\begin{itemize}
    \item openai~=~1.35.10
    \item graphviz~=~0.20.3
    \item requests~=~2.32.3
    \item wikipedia~=~1.4.0
    \item tqdm~=~4.66.4
    \item termcolor~=~2.4.0
    \item fastapi~=~0.111.1
    \item matplotlib~=~3.9.1
    \item faiss-cpu~=~1.8.0.post1
    \item loguru~=~0.7.2
    \item pytest-asyncio~=~0.24.0
    \item websockets~=~13.0.1
    \item transformers~=~4.44.2
    \item torch~=~2.4.1
\end{itemize}

A system with at least 32GB RAM and a GPU for faster LLM inference is recommended.

The system relies on the \texttt{AsyncHow-Based Agentic Systems Evaluation Dataset} available at \href{https://anonymous.4open.science/r/AsyncHow-Based-Agentic-Systems-Evaluation-Dataset-C6BD/README.md}{link}.

\subsubsection{Core Components and Function Shells}

The \textbf{Orchestrator} is responsible for generating a Directed Acyclic Graph (DAG) from user queries, breaking them into manageable tasks.

\begin{verbatim}
class Orchestrator:
    def __init__(self, gpt_client_manager):
        """
        Initializes the Orchestrator with the GPT client manager for 
        processing user queries into a task graph (DAG).
        """
        self.client = gpt_client_manager

    def produce_task_graph(self, user_query):
        """
        Calls the LLM to convert a user query into a task graph.
        The task graph is a DAG representing tasks 
        and their dependencies.
        """
        return task_graph
\end{verbatim}

The \textbf{ToolManager} dynamically loads Python functions from files and selects relevant tools for each task.

\begin{verbatim}
class ToolManager:
    def __init__(self):
        """
        Initializes the ToolManager, which is 
        responsible for loading, embedding, and 
        selecting tools for execution 
        based on task descriptions.
        """

    def parse_all_function_files(self, 
                        function_files_path="python_tools"):
        """
        Parses Python files containing 
        tool functions and embeds their descriptions
        to make them available for 
        selection during task execution.
        """

    def filter_tools_by_tasks(self, task_list):
        """
        Filters available tools using 
        FAISS based on the semantic similarity 
        between the task description and tool embeddings.
        """
\end{verbatim}

The \texttt{ToolManager} links directly to the \texttt{GraphExecutor}, which uses filtered tools for task execution.

The \textbf{GraphExecutor} handles task execution, using tools selected by the ToolManager. It can execute tasks in parallel or sequentially, depending on dependencies.

\begin{verbatim}
class GraphExecutor:
    def __init__(self, my_gpt_client_manager,
                    include_indirect_dependencies=False,
                    timing_profiler=None):
        """
        Initializes the GraphExecutor, 
        which executes tasks either sequentially or
        in parallel while handling dependencies.
        """
        self.client = my_gpt_client_manager.openai_client
        self.tool_manager = None
        self.timing_profiler = timing_profiler

    def initialize_tool_manager(self, tool_manager):
        """
        Sets the ToolManager instance, which 
        will be used to filter and select tools
        for task execution.
        """

    def execute_task_graph_sequentially(self):
        """
        Executes tasks in the task graph sequentially, 
        respecting the dependencies between tasks.
        """
        return execution_results

    def execute_task_graph_parallely(self):
        """
        Executes tasks in the task graph in 
        parallel where possible, 
        optimizing for concurrency.
        """
        return execution_results

    def _execute_task_func(self, task, inter_task_buffer):
        """
        Executes a single task, interacting 
        with the GPT model for task results,
        and managing the inter-task message buffer 
        to handle dependencies.
        """
        return final_response, execution_timing, tool_calls
\end{verbatim}

The \texttt{GraphExecutor} interacts with the \texttt{ToolManager} to call relevant tools and with the \texttt{Delegator} for managing task execution order.

The \textbf{Delegator} manages the execution of tasks by ensuring proper sequencing and handling inter-task communication.

\begin{verbatim}
class Delegator:
    def __init__(self, pipeline):
        """
        Initializes the Delegator, which is 
        responsible for managing the flow of
        tasks, including communication between them.
        """
        self.pipeline = pipeline

    def execute_task_graph(self):
        """
        Orchestrates the execution of the entire task graph,
        ensuring each task is executed in the 
        correct order and managing inter-task buffers.
        """
        return task_results
\end{verbatim}

The \texttt{Delegator} consolidates results from the \texttt{GraphExecutor} and ensures that tasks dependent on others receive the necessary input.

The \textbf{Feedback System} generates real-time feedback for users during task execution, keeping them informed about task progress.

\begin{verbatim}
class FeedbackSystem:
    def add_to_queue(self, task_id, task_label):
        """
        Adds a task to the feedback queue 
        to provide real-time updates to users
        about the progress of that specific task.
        """
\end{verbatim}

The \texttt{FeedbackSystem} is integrated with the \texttt{GraphExecutor} to deliver updates to the user as tasks are executed.

The \textbf{Timing Profiler} tracks execution times for tasks and tools, enabling detailed performance analysis.

\begin{verbatim}
class TimingProfiler:
    def start_task_timing(self, task_id, task_label):
        """
        Starts tracking the execution time 
        for a specific task, storing the start time.
        """

    def stop_task_timing(self, task_id):
        """
        Stops tracking the execution time 
        for a specific task, storing the
        end time and calculating the duration.
        """
\end{verbatim}

The \texttt{TimingProfiler} links to both the \texttt{GraphExecutor} and the \texttt{FeedbackSystem} for detailed time tracking of task execution and feedback generation.

The \textbf{Pipeline} class orchestrates the entire framework, ensuring proper initialization, task decomposition, and execution. It interacts with the Orchestrator, ToolManager, GraphExecutor, and FeedbackSystem to process user queries.

\begin{verbatim}
class Pipeline:
    def __init__(self, semantic_tool_filtering=True, 
                 include_indirect_dependencies=True, 
                 generate_feedback=True, 
                 profile_execution_timings=True):
        """
        Initializes the Pipeline with various options, including:
        - Semantic tool filtering: 
            Filters tools based on task relevance.
        - Include indirect dependencies: 
            Includes results from all ancestor tasks.
        - Generate feedback: 
            Provides real-time feedback during task execution.
        - Profile execution timings: 
            Tracks the execution time for tasks and tools.
        """
        self.semantic_tool_filtering = semantic_tool_filtering
        self.include_indirect_dependencies = include_indirect_dependencies
        self.generate_feedback = generate_feedback
        self.profile_execution_timings = profile_execution_timings
        self.tool_manager = ToolManager()
        self.graph_executor = None
        self.orchestrator = Orchestrator(gpt_client_manager)
        self.feedback_system = None

    def configure_for_query(self, user_query):
        """
        Configures the pipeline for a new user query. 
        This includes initializing the task graph, 
        selecting tools, and setting up 
        feedback and profiling options.
        """
        task_graph = self.orchestrator.produce_task_graph(user_query)
        self.graph_executor = GraphExecutor(self.orchestrator.client, 
                                        self.include_indirect_dependencies, 
                                        self.profile_execution_timings)
        self.graph_executor.initialize_tool_manager(self.tool_manager)
        self.graph_executor.initialize_task_graph(task_graph)

    def run_with_pretty_print(self, user_query):
        """
        Executes the user query by 
        first configuring the pipeline and then running
        the task graph execution either 
        sequentially or in parallel. 
        Outputs are displayed with real-time feedback 
        and execution timings.
        """
        self.configure_for_query(user_query)
        if self.generate_feedback:
            self.graph_executor.initialize_feedback_system(
                                        self.feedback_system)

        # Executes the task graph and provides pretty-printed results
        results = self.graph_executor.execute_task_graph_sequentially()  
        # or parallely
        print(results)
\end{verbatim}

The \texttt{Pipeline} class links directly to all the major components (\texttt{Orchestrator}, \texttt{ToolManager}, \texttt{GraphExecutor}, and \texttt{FeedbackSystem}) and acts as the main interface for the entire framework. It ensures that tasks are processed, tools are selected, and feedback is generated in a streamlined way.

\subsubsection{Prompt Skeletons}
The framework uses specific prompts to interact with the LLM (GPT-4). These prompts handle task graph generation, task execution, tool selection, and user feedback. Below is a skeleton of the main prompts from the system.

\begin{verbatim}
PROMPT_TASK_GRAPH = """
You are responsible for generating a task graph 
from the following user query. Decompose the query 
into individual tasks and create a Directed Acyclic Graph (DAG) 
with nodes as tasks and edges as dependencies. 
Ensure there are no cyclic dependencies.

User Query: {user_query}

Respond with the task graph in the following JSON format:
{
  "nodes": [
    {"id": 1, "label": "Task description 1"},
    {"id": 2, "label": "Task description 2"}
  ],
  "edges": [
    {"from": 1, "to": 2}
  ]
}
"""

PROMPT_CONSOLIDATE_TASKS = """
You are an assistant operating within an LLM-based Agentic Architecture. 
Your task is to generate a final response to the user's query 
by considering the results of multiple tasks. 
These tasks were generated from the user's query 
using a task graph. Ensure the final response 
addresses all aspects of the user's query.

User Query: {user_query}

Task Results: {task_results}

Generate a concise final response in 50 words or less.
"""

PROMPT_FEEDBACK = """
You are responsible for generating a short feedback phrase 
for a task that is being processed. 
The feedback should be friendly and let the user 
know their task is in progress.

Task: {task_description}

Generate a new feedback phrase:
"""
\end{verbatim}

\texttt{PROMPT\_TASK\_GRAPH}: This prompt is used to decompose a user query into a Directed Acyclic Graph (DAG). Each node represents a task and edges represent dependencies.

\texttt{PROMPT\_CONSOLIDATE\_TASKS}: This prompt consolidates the results of multiple tasks into a single final response to the user's query.
 
 \texttt{PROMPT\_FEEDBACK}: This prompt generates feedback to inform the user of task progress during execution.

\subsubsection{Replication Guide}

\textbf{Step 1: Environment Setup}
Install dependencies using the provided \texttt{requirements.txt} file:
\begin{verbatim}
pip install -r requirements.txt
\end{verbatim}

\textbf{Step 2: Configure GPT-4}
Set up GPT-4 through an Azure OpenAI deployment and configure credentials in \texttt{gpt\_client.py}.

\textbf{Step 3: Prepare Tools}
Add your custom Python tools in the \texttt{python\_tools} folder. The \texttt{ToolManager} will automatically detect and embed these tools.

\textbf{Step 4: Execute Pipeline}
Run the pipeline with a sample user query:
\begin{verbatim}
query = "What time does the sun rise today?"
pipeline.run_with_pretty_print(query)
\end{verbatim}

\textbf{Step 5: Analyze Results}
The execution results and timing profile will be stored in a JSON file for performance analysis.

This guide, with function shells, descriptions, step-by-step instructions and prompt skeletons, provides a detailed replication process for the agentic framework. The modular nature of the components allows for easy customization and adaptation to other domains.

\subsection{Evaluation and Dataset Creation}\label{supplementary_replicate_evaluation}

To evaluate the performance of our agentic task decomposition framework, we generated a custom evaluation dataset. This section describes the process of creating task graphs, tools, and the evaluation metrics used for analyzing the framework.

\subsubsection{Dataset Creation}
The evaluation dataset is built by creating task graphs for sequential, parallel, and asynchronous task execution scenarios. Each scenario is associated with a list of tools, and the task graph is generated based on tool descriptions.

\begin{verbatim}
def create_seq_task_graph(edges, node_descriptions):
    """
    Creates a sequential task graph 
    based on a list of edges and node descriptions.
    This function removes 'Start' and 'End' nodes 
    and maps tasks to unique IDs.
    
    Args:
        edges (list of tuples): List of edges 
        where each edge is a tuple (from_node, to_node).
        node_descriptions (list of str): List of 
        task descriptions.

    Returns:
        dict: A dictionary representing the task graph 
        with nodes and edges.
    """
    task_map = {}
    node_list = []
    current_task_id = 1

    for i, description in enumerate(node_descriptions):
        if description not in ["Start", "End"]:
            task_id = f"task_{current_task_id}"
            task_map[str(i+1)] = task_id
            node_list.append({'id': task_id, 
                              'label': description})
            current_task_id += 1

    edge_list = []
    for from_node, to_node in edges:
        if from_node in task_map and to_node in task_map:
            edge_list.append({'from': task_map[from_node], 
                              'to': task_map[to_node]})

    return {'task_graph': {'nodes': node_list, 
                           'edges': edge_list}}
\end{verbatim}

This function is responsible for creating task graphs from a list of task descriptions and edges. A similar function is used for generating task graphs for parallel tasks.

\begin{verbatim}
def create_parallel_graph(edges, tools):
    """
    Creates a parallel task graph based on a list of tools. No edges are added.
    
    Args:
        edges (list of tuples): List of edges 
        representing the dependencies (if any).
        tools (list of str): List of tool names.

    Returns:
        dict: A dictionary representing the task graph with nodes and edges.
    """
    task_graph = {
        "task_graph": {
            "nodes": [],
            "edges": []
        }
    }

    for i, tool in enumerate(tools):
        task_id = i + 1
        task_graph["task_graph"]["nodes"].append({"id": task_id, 
                                                  "label": tool})

    return task_graph
\end{verbatim}

\subsubsection{Evaluation Process}

We evaluated the task decomposition framework using synthetic tasks designed for both sequential and parallel task execution. The following function helps create a structured evaluation dataset:

\begin{verbatim}
def create_async_graph(edges, node_descriptions):
    """
    Creates an asynchronous task graph by mapping nodes to unique integer IDs.
    
    Args:
        edges (list of tuples): List of edges 
        where each edge is a tuple (from_node, to_node).
        node_descriptions (list of str): List of 
        task descriptions.

    Returns:
        dict: A dictionary representing the task graph 
        with nodes and edges.
    """
    task_graph = {
        "task_graph": {
            "nodes": [],
            "edges": []
        }
    }

    for i, desc in enumerate(node_descriptions):
        task_graph["task_graph"]["nodes"].append({"id": i+1, 
                                                  "label": desc})

    for from_node, to_node in edges:
        task_graph["task_graph"]["edges"].append({"from": from_node, 
                                                  "to": to_node})

    return task_graph
\end{verbatim}

During the dataset creation, semantic duplicates are removed to ensure a high-quality dataset.

\begin{verbatim}
def remove_semantic_duplicates(folder_path, 
                               similarity_threshold=0.8):
    """
    Identifies and removes semantically similar 
    duplicates from the folder names.
    
    Args:
        folder_path (str): Path to the folder 
        containing subfolders.
        similarity_threshold (float): Threshold above which 
        two folders are considered duplicates.

    Returns:
        list: A list of unique subfolder names.
    """
    model = SentenceTransformer('all-MiniLM-L6-v2')
    subfolder_names = [name for name in os.listdir(folder_path) 
    if os.path.isdir(os.path.join(folder_path, name))]
    embeddings = model.encode(subfolder_names, 
                              convert_to_tensor=True)

    unique_subfolders = []
    for i, name in enumerate(subfolder_names):
        if not any(util.cos_sim(embeddings[i], embeddings[j]) > 
        similarity_threshold for j in range(i)):
            unique_subfolders.append(name)

    return unique_subfolders
\end{verbatim}

Tools for the evaluation were synthetically generated to mimic real-world APIs or data retrieval functions. Below is the prompt-based function for creating these tools:

\begin{verbatim}
def generate_functions(data, folder):
    """
    Generates Python functions based on 
    the tool descriptions and stores 
    them in the specified folder.
    
    Args:
        data (dict): Contains tool names and 
        corresponding function code.
        folder (str): Path to the folder where the 
        functions will be saved.
    """
    base_directory = os.path.join("../..", folder)

    for function_name, function_code in zip(data["function_names"], 
                                            data["functions"]):
        function_directory = os.path.join(base_directory, 
                                          function_name)
        os.makedirs(function_directory, exist_ok=True)

        file_path = os.path.join(function_directory, 
                                f"{function_name}.py")
        with open(file_path, "w") as file:
            file.write(function_code)
\end{verbatim}

\subsubsection{Prompt Skeletons}

Below are the prompt skeletons used during the evaluation and dataset creation process. These prompts guide the LLM in generating Python functions, removing duplicates, and consolidating tool outputs.

\begin{verbatim}
PROMPT_GENERATE_FUNCTIONS = """
This prompt instructs the LLM to generate Python functions 
based on a list of tool descriptions. 
It includes instructions for creating synthetic data 
and formatting the function signatures and bodies.

### Input:
- Tool Descriptions: {tools}
- User Query: {query}

### Output:
- JSON object containing the functions, 
their names, and a gold standard response.
"""
\end{verbatim}

\begin{verbatim}
PROMPT_REMOVE_DUPLICATES = """
This prompt guides the system in identifying and 
removing semantically similar duplicates 
from a dataset folder. It ensures that no 
two tools or tasks are too similar.

### Input:
- Folder Path: {folder_path}
- Similarity Threshold: {similarity_threshold}

### Output:
- List of unique tool/task names.
"""
\end{verbatim}

\begin{verbatim}
PROMPT_CONSOLIDATE_TASKS = """
This prompt consolidates the results of several tool 
executions into a final response. 
It ensures that the final output is concise, accurate, 
and addresses the user's query.

### Input:
- Task Results: {task_results}

### Output:
- Consolidated response based on tool outputs.
"""
\end{verbatim}

\begin{verbatim}
PROMPT_COMPLEXITY_SCORE = """
This prompt categorizes user queries based on their complexity, 
assigning them into predefined categories. 
It ensures that each query is analyzed for its difficulty and 
aligned with the appropriate evaluation metric.

### Input:
- User Query: {query}

### Output:
- A complexity score/category for the query.
"""
\end{verbatim}

\texttt{PROMPT\_GENERATE\_FUNCTIONS}: Generates Python functions based on tool descriptions, ensuring realistic outputs and proper formatting.

\texttt{PROMPT\_REMOVE\_DUPLICATES}: Helps identify and remove semantically similar duplicates from a dataset.

\texttt{PROMPT\_CONSOLIDATE\_TASKS}: Consolidates tool execution results into a single, accurate response.

\texttt{PROMPT\_COMPLEXITY\_SCORE}: Categorizes user queries into complexity levels for evaluation purposes.

\subsubsection{Replication Guide}

To replicate the dataset creation and evaluation process, follow these steps:

\textbf{Step 1: Setup the Environment}
Install the necessary dependencies and tools listed in the \texttt{requirements.txt} file. Ensure that the system has access to an LLM such as GPT-4 through the OpenAI API or Azure OpenAI.

\begin{verbatim}
pip install -r requirements.txt
\end{verbatim}

\textbf{Step 2: Generate Task Graphs}
Run the provided Python scripts to generate task graphs for sequential, parallel, and asynchronous scenarios. These graphs will form the backbone of the evaluation dataset, mapping tasks to tools and establishing dependencies.

\begin{verbatim}
# Generate sequential task graphs
create_seq_graphs(df_name="asynchow_seq_df.csv")

# Generate parallel task graphs
create_parallel_graphs(df_name="asynchow_para_df.csv")

# Generate asynchronous task graphs
create_async_graphs(df_name="asynchow_async_df.csv")
\end{verbatim}

\textbf{Step 3: Generate Python Functions for Tools}
For each tool in the dataset, generate a Python function that mimics its behavior. Use the function generation script to create these tools and save them in the designated folder. This step involves feeding tool descriptions to the LLM, which generates the corresponding Python code.

\begin{verbatim}
# Generate functions based on tool descriptions
generate_functions_from_tool_list()
\end{verbatim}

\textbf{Step 4: Remove Semantic Duplicates}
Ensure that no redundant or semantically similar tools are included in the dataset by running the duplicate removal script.

\begin{verbatim}
remove_semantic_duplicates(folder_path="tools", similarity_threshold=0.8)
\end{verbatim}

\textbf{Step 5: Execute and Evaluate}
Finally, execute the tasks using the generated tools and evaluate the system's performance by analyzing the task results and feedback.

\begin{verbatim}
# Run the task execution and feedback generation
pipeline.run_with_pretty_print(query)
\end{verbatim}

After running the evaluation, analyze the complexity scores and tool performance to measure the framework’s ability to handle complex task decomposition and execution workflows.

This process generates a complete evaluation dataset, tests the system across multiple task execution scenarios, and provides detailed insights into the system's performance. By following the outlined steps, researchers and developers can replicate the evaluation process and extend the framework to additional scenarios or datasets.

\newpage
\appendix
\section*{Appendix}
\label{Appendix_Data_Generation}
\section{Evaluation Plots} \label{appendix_eval_plots}
The plots show the statistically significant correlations in regards to the Answer Score.

\begin{figure}[h]
\centering
\includegraphics[width=0.8\textwidth]{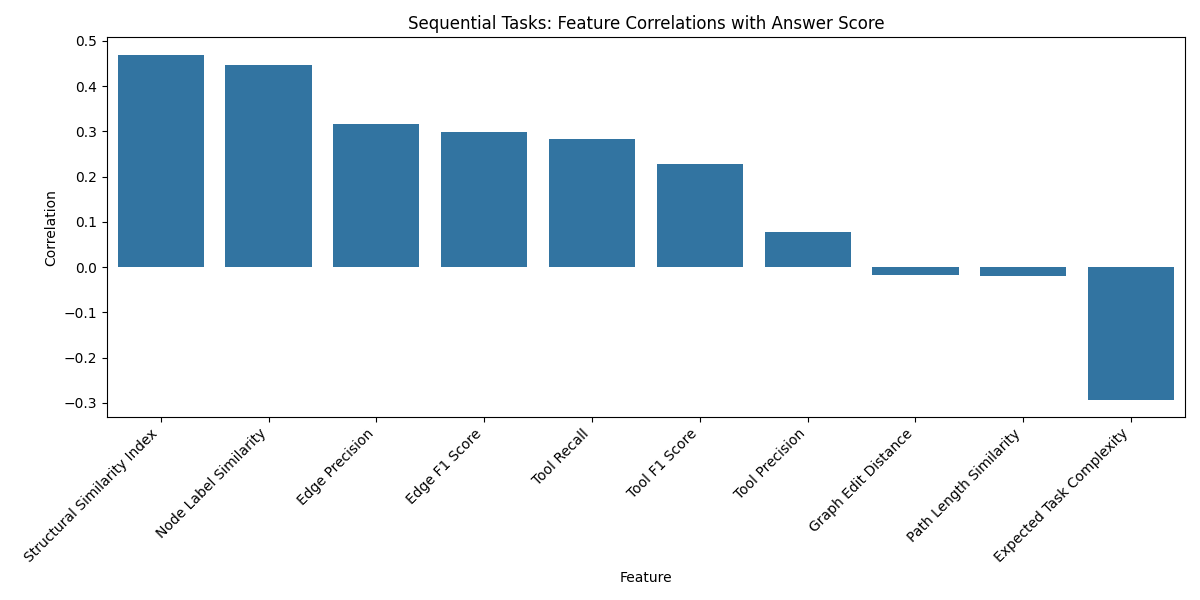}
\caption{Feature Correlations with Answer Score for Sequential Tasks}
\label{fig:seq_correlations}
\end{figure}

\begin{figure}[h]
\centering
\includegraphics[width=0.8\textwidth]{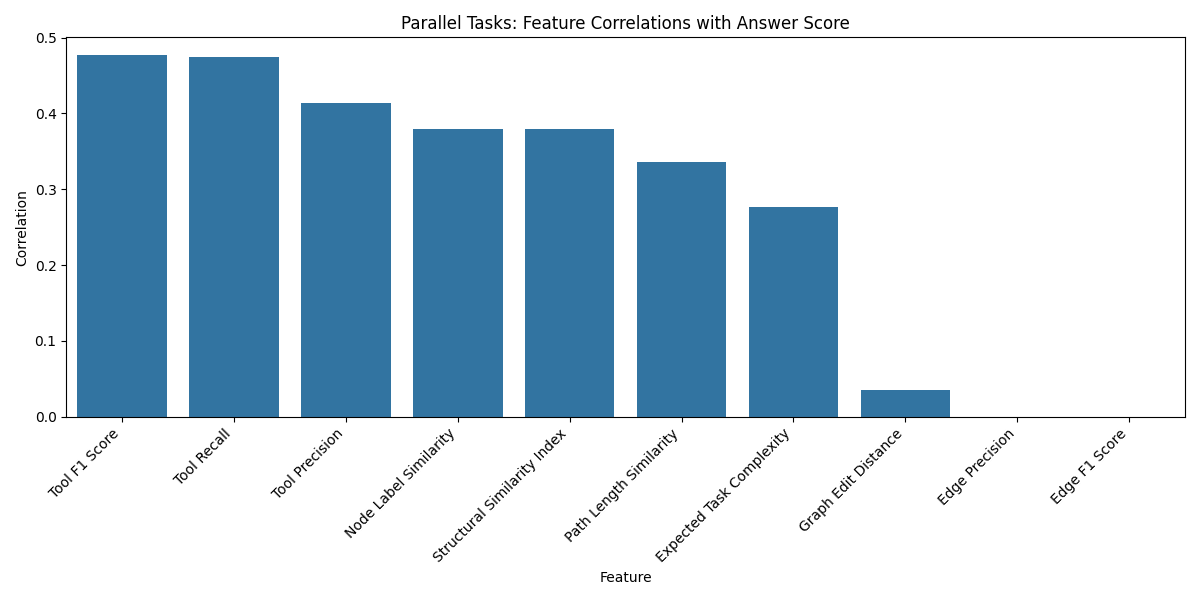}
\caption{Feature Correlations with Answer Score for Parallel Tasks}
\label{fig:para_correlations}
\end{figure}

\section{Additional plots for timing profiling of parallel execution of task graphs}\label{appendix_additional_executor_plots}

\begin{figure}[h]
    \centering
    \includegraphics[width=0.5\linewidth]{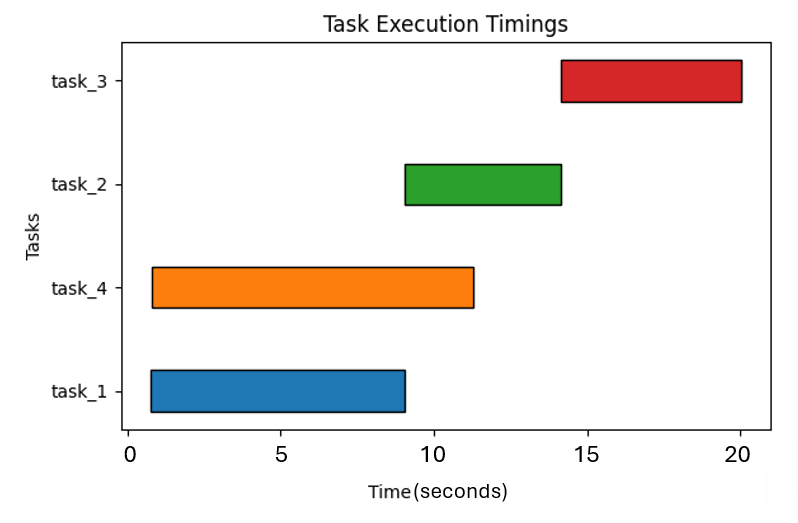}
    \caption{Task execution timings for parallel execution. The diagram shows the start and end times for four tasks (Task 1, Task 2, Task 3, and Task 4). Tasks 1 and 4, which are independent, are executed in parallel, while Task 2 starts after Task 4, and Task 3 follows Task 2, showing sequential execution due to dependencies.}
    \label{fig:execution-timings}
\end{figure}

Figure \ref{fig:execution-timings} shows the specific timing of tasks during a parallel execution scenario. This timing diagram demonstrates how independent tasks (e.g., Task 1 and Task 4) are executed in parallel, while tasks that have dependencies (e.g., Task 2 and Task 3) are scheduled sequentially, respecting their dependencies. The timing diagram demonstrates the framework’s ability to efficiently manage parallel execution while ensuring dependent tasks are executed in the correct order.

\section{Additional Future Work}
\label{appendix_additional_future_work}
Based on the results and implications of our current study, we plan to explore several avenues for future work that could further enhance the capabilities and evaluation of agentic systems:

\begin{itemize}
    \item \textbf{Multi-Agent Communication:} Developing and integrating multi-agent communication capabilities to enhance the collaborative aspect of our framework. This would allow for more complex task execution scenarios where multiple agents must coordinate and share information.

    \item \textbf{Causal Influence Score (CIS):} Developing the CIS metric to evaluate the causal relationships between the tools used and the final outcomes or answers generated by the agent. Unlike traditional precision and recall metrics, CIS would focus on understanding how each tool's use causally impacts overall task success.

    \item \textbf{Dynamic Adaptation Metric (DAM):} Introducing the DAM metric to measure an agent’s ability to dynamically adapt its tool selection and task execution strategies in response to changing conditions or feedback during task execution.

    \item \textbf{Inter-tool Dependency Metric (ITDM):} Creating the ITDM metric to evaluate the degree of dependency between tools used by the agent within a task. This metric would capture whether the agent effectively coordinates between multiple tools or relies too heavily on certain tools.

    \item \textbf{Task Graph Robustness Index (TGRI):} Implementing the TGRI metric to measure the robustness of the task graph generated by the agent against errors or changes. This metric would evaluate how small modifications to the input or environment affect the task graph and its resulting accuracy.

    \item \textbf{Contextual Consistency Score (CCS):} Developing the CCS metric to measure how consistently an agent applies tools and constructs task graphs in contexts similar to previously encountered scenarios. This would assess the agent’s ability to generalize across similar tasks.

    \item \textbf{Explainability Score (ES):} Introducing the ES metric to measure how well the agent can explain its decisions, particularly its choice of tools and task graph construction. This metric would be crucial for ensuring transparency and building trust in agentic systems.

    \item \textbf{Cumulative Learning Rate (CLR):} Developing the CLR metric to measure how quickly and effectively the agent improves its performance over time as it encounters more tasks. This would capture the agent's learning efficiency and its ability to adapt and improve.

    \item \textbf{Interaction Cost Metric (ICM):} Creating the ICM metric to evaluate the efficiency of an agent's interactions in terms of resource usage, such as time, computational power, or energy, in achieving a task. This metric would be vital for optimizing the resource efficiency of agentic systems.
\end{itemize}

These future directions aim to build on our current findings, enhancing both the theoretical and practical aspects of agentic system design and evaluation. By developing these additional metrics and capabilities, we hope to provide a more comprehensive framework for assessing and improving the performance of agentic systems in increasingly complex environments.

\end{document}